\documentclass[11pt,a4paper]{article}
\usepackage{authblk}

\usepackage[hyperref]{acl2021}
\usepackage{times}
\usepackage{latexsym}

\usepackage{times}
\usepackage{latexsym}
\usepackage{pbox}
\usepackage{makecell}
\usepackage{latexsym}
\usepackage{graphicx}
\usepackage{subcaption}
\usepackage{setspace}
\usepackage{enumitem}
\usepackage{booktabs, cellspace, multirow}
\usepackage{amsmath}
\usepackage{amssymb}

\usepackage{xcolor}

\definecolor{myblue}{rgb}{0.12, 0.47, 0.71}
\definecolor{mygreen}{rgb}{0.11, 0.62, 0.47}
\definecolor{myorange}{rgb}{0.85, 0.37, 0.01}
\definecolor{myred}{rgb}{0.84, 0.15, 0.16}
\definecolor{mypurple}{rgb}{0.58, 0.40, 0.74}
\usepackage{microtype}
\newcommand*{\skippingparagraph}{\par\vspace{0.3em}\noindent}

\usepackage{xcolor}

\aclfinalcopy 
\def\aclpaperid{3086} 

\title{Synthesizing Adversarial Negative Responses for Robust Response Ranking and Evaluation}

\author{Prakhar Gupta$^\clubsuit$ \quad Yulia Tsvetkov$^\spadesuit$ \quad Jeffrey P. Bigham$^{\clubsuit,\heartsuit}$  \\
$^\clubsuit$Language Technologies Institute, Carnegie Mellon University \\
$^\spadesuit$Paul G.~Allen School of Computer Science \& Engineering, University of Washington \\
$^\heartsuit$Human-Computer Interaction Institute, Carnegie Mellon University \\
\texttt{\small prakharg@cs.cmu.edu, yuliats@cs.washington.edu, jbigham@cs.cmu.edu}}
\date{}

\begin{document}
\maketitle
\begin{abstract}
Open-domain neural dialogue models have achieved high performance in response ranking and evaluation tasks.
These tasks are formulated as a binary classification of responses given in a dialogue context, and models generally learn to make predictions based on context-response content similarity. However, over-reliance on content similarity makes the models less sensitive to the presence of inconsistencies, incorrect time expressions and other factors important for response appropriateness and coherence.
We propose approaches for automatically creating adversarial negative training data to help ranking and evaluation models learn features beyond content similarity.
We propose mask-and-fill and keyword-guided approaches that generate negative examples for training more robust dialogue systems. These generated adversarial responses have high content similarity with the contexts but are either incoherent, inappropriate or not fluent.
Our approaches are fully data-driven and can be easily incorporated in existing models and datasets. Experiments on classification, ranking and evaluation tasks across multiple datasets demonstrate that our approaches outperform strong baselines in providing informative negative examples for training dialogue systems.\footnote{ Code and data are publicly available \url{https://github.com/prakharguptaz/Adv\_gen\_dialogue}}

\end{abstract}

\section{Introduction}
\label{introduction}
Due to growing availability of dialogue corpora~\cite{li-etal-2017-dailydialog, zhang-etal-2018-personalizing, smith-etal-2020-put} and the advancement of neural architectures~\cite{Radford2019LanguageMA,brown2020language, devlin-etal-2019-bert}, dialogue systems have achieved considerable success.
As typically formulated, dialogue models generate one or more candidate responses to a provided context, consisting of past dialogue turns.
Dialogue ranking~\cite{zhou-etal-2018-multi, wu-etal-2019-sequential} and evaluation models~\cite{ruberbib,yi-etal-2019-towards,sato-etal-2020-evaluating}, in turn, are deployed to select and score candidate responses according to coherence and appropriateness. 

Ranking and evaluation models are generally trained using true positive responses and randomly selected negative responses, which raises
two issues. 
First, random negative candidates often have low content similarity with the context, and thus models learn to associate response coherence and appropriateness with content similarity~\cite{ yuan-etal-2019-multi, whang2020response, sai2020improving}. In real systems, generated response candidates tend to be more similar in terms of content, and so other factors (e.g.,~time expressions, dialogue acts, inconsistencies) tend to be more important.
Second, randomly selecting candidates as negative examples in an open domain context can result in false negatives,
leading to misclassification of appropriate responses.

\begin{table*}[t]
    \small
    \begin{tabular}{  l  p{2.2cm} p{4.1cm}  p{6.6cm} }
        \toprule
      
& \textbf{Error category}
& \textbf{Description}
& \textbf{Sample responses} \\\midrule

\textbf{C-ent}
    & Incorrect entities or actors (R,G)
    & Incorrect subject or object of verbs or presence of one or more incorrect entities or coreference.
    & \textit{Context:} I am so happy that you are doing okay. \newline
    \textit{Response:} My friend is always happy.
    \\\hline

\textbf{C-time}
    & Incorrect Time expressions (R)        
    & Use of incorrect time expressions or tense of verbs.
  & \textit{Context:} What are you going to do on Monday? \newline
    \textit{Response:} Yesterday, I celebrated my daughter's wedding anniversary.
        \\\hline    

\textbf{C-cont}
    & Contradictory or extraneous details (R,G)     
    & Presence of details which make the response inconsistent within itself or contradict the context
    & \textit{Context:} A: I don't know why I bothered to come here. 
    \newline
    B: Did you enjoy your stay? \newline
    \textit{Response:} I enjoyed the concert a lot.
    \\\hline 
\textbf{C-speaker}
    & Incorrect speaker turn (R) 
    & The response is relevant to the conversation but from the wrong speaker.
    &   \textit{Context:} What starting salary would you expect here? \newline
    \textit{Response:} If you work overtime, I will pay you extra salary.
    \\\hline

\textbf{C-follow}
    & Does not directly address the context (R,G) 
    & The response does not follow immediately from the context.
    & \textit{Context:}  What would you like for main course sir? \newline
    \textit{Response:} I know very well how to make noodles, and I taught one of my friends.
    \\\hline    

\textbf{C-strat}

    & Incorrect strategies (R,G)     
    & Use of incorrect dialogue act, emotion, persona or style
    & \textit{Context:} I can't find the paper clips. \newline
    \textit{Response:} Ok, great work.
        \\\hline

\textbf{C-lang}

    & Poor language (G)       
    & Presence of poor grammar, incorrect sentence structures or repetitions
    & \textit{Context:} Do you have mixed drinks available here? \newline
    \textit{Response:} Yes. This order is divided by 16 divided for main main ones of order.\\
    \bottomrule
    \end{tabular}
    \caption{Error categories prevalent in inappropriate responses with high context-response semantic relatedness. We present 7 categories with their descriptions and sample context and response pairs. For each category we also indicate whether it is frequently observed in Retrieval (R) or Generation (G) models. Models which simply learn to associate response coherence with content similarity often ignore these errors. Our approaches create adversarial negative data for training dialogue models by introducing such errors in context relevant utterances.}
    \vspace{-0.5pc}
    \label{tab:categories}
\end{table*}

To make dialogue models more robust to the spurious pattern of content similarity, prior work proposed to leverage adversarial and counterfactual examples~\cite{kaushik2019learning, srivastava2020robustness}.
A reliable method for creating counterfactual data is to collect  human-written adversarial negative responses
\cite{sai2020improving}, but it is 
expensive, time-consuming, and difficult to scale. Our goal is to create reliable automatic methods for \emph{synthesizing} adversarial negative responses.

The most common approach to generating natural language adversarial examples is to paraphrase or insert typos, synonyms, or words relevant to the context in the inputs~\cite{iyyer-etal-2018-adversarial,ebrahimi-etal-2018-hotflip,alzantot-etal-2018-generating,zhang-etal-2019-generating-fluent}.
In open domain conversations, however, a context can have a wide range of possible responses with varied forms and semantics.
Small lexical variations via substitutions and paraphrasing do not provide adequate coverage over the possible space of adversarial responses, and they can also lead to generation of false negatives due to the open-ended nature of dialogues.
Creating adversarial dialogue responses is thus different, and can be more challenging than in other natural language domains.

We propose two approaches for adversarial response creation: 1) a mask-and-fill approach that corrupts gold responses related to the context but retains content similarity, and 2) a keyword-guided generative approach that uses concepts from the context to generate topically relevant but incoherent responses. These approaches do not require additional annotations, are black-box (do not need access to model parameters), and are easily adapted to new datasets and domains.

The main contributions of this paper are: 1) We identify and discuss error patterns present in retrieval and generation model outputs, which are difficult to detect due to high content similarity; 
2) To the best of our knowledge, we are the first to propose automatic approaches for creating adversarial responses for dialogue model training in a black-box setting; and,
3) We demonstrate that our proposed approaches achieve better performance compared to strong baselines on two datasets on dialogue classification, ranking and evaluation tasks.

\section{Properties of Adversarial Responses}
\label{sec:properties}
\vspace{-2pt}
Models trained using randomly sampled negative examples tend to assign high scores to responses with high content similarity with the context, and often ignore other important factors necessary for response appropriateness and coherence.
Therefore, we aim to generate adversarial negative responses which have high content similarity with the context, but which still possess factors rendering the responses inappropriate to the context. 
We present the categorization of such factors or error types which can make a response inappropriate in Table~\ref{tab:categories}. For each category, we provide its description and sample context-response pairs. 
To create this categorization, we manually analyzed responses present in outputs of generative models, candidates of retrieval sets, and human written adversarial dialogue responses~\cite{sai2020improving}.
Categories C-ent, C-time and C-cont are errors related to various inconsistencies and logical flaws in the responses and indicate poor response \textit{appropriateness}. Categories C-speaker, C-follow and C-strat are error types specific to the dialogue setting and indicate poor response \textit{coherence}. Category C-lang indicates poor response \textit{fluency}. Our categorization of errors is inspired by the categorization suggested by~\citet{pagnoni-2021-frank} for factuality of summarization, and \citet{higashinaka2019improving, ko-etal-2019-linguistically} and \citet{ sato-etal-2020-evaluating} for dialogue. These categories inform our approaches as well as error analysis.

\section{Methodology}
For a given dialogue context $C$ and its gold response \textit{$R_g$}, our goal is to generate an adversarial response $R_a$ such that while achieving high scores from dialogue ranking or evaluation models, it should not be a valid response to the context $C$. 
Dialogue ranking and evaluation models trained with such hard synthetic negative responses should learn to associate response relevance with features beyond content similarity, and hence become robust against spurious features.


The adversarial responses should satisfy the following criteria: 1) have high content similarity with input contexts; 2) have one or more errors (Table~\ref{tab:categories}) which make the response inappropriate to the context; 3) be hard training examples, that is, they should likely be misclassified by current models as correct; and 4) sufficiently cover errors which occur naturally in model generated responses and retrieval candidates, and therefore they should be plausible and diverse.
We propose two approaches for synthesizing adversarial negative examples - a mask-and-fill approach and a keyword-guided generation approach which we discuss next.

\subsection{Mask-and-fill Approach}
\label{sec:masknfill}
\begin{figure}[tb]
    \centering
    \vspace{-.2pc}
    \includegraphics[width=0.46\textwidth]{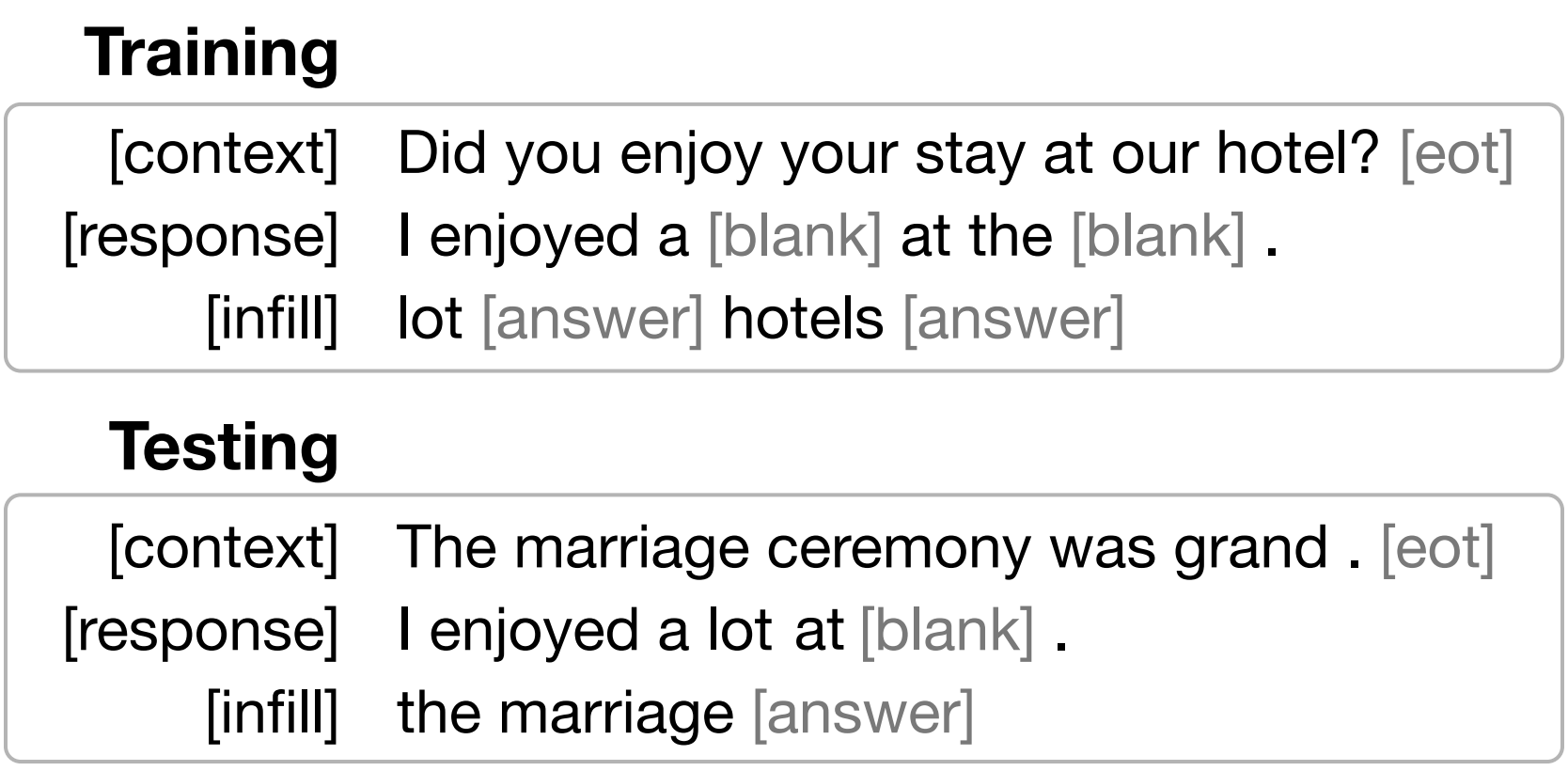}
    \vspace{-.5pc}
    \caption{\textit{Mask-and-fill} approach using ILM model. ILM is trained to infill n-grams in place of blanks in a response. Tokens after [infill] replace the [blank] tokens. During training, \textit{Mask-and-fill} learns to infill responses conditioned on the correct context. During testing, it infills the response conditioned on a random context which introduces errors in the response.}
    \label{fig:ilm}
     \vspace{-1em}
\end{figure}

This approach modifies and corrupts original utterances related to a context as shown in Figure~\ref{fig:ilm}. It consists of two steps: 1) masking, where one or more tokens of an original utterance are masked out; and 2) infilling, where the masked out tokens are substituted with new tokens.
For a context \textit{$C$}, the set of original utterances consists of:
\begin{itemize}[topsep=1pt,itemsep=0pt,partopsep=0pt, parsep=0pt,leftmargin=*]
\item Set of ground truth responses of the context - $R_g$. 
\item Set of utterances from the context - $U_c$.
\item Set of retrieved responses based on context - \textit{$R_e$}.
\end{itemize}

\noindent
\textbf{Masking}: We use the hierarchical masking function from \citet{donahue-etal-2020-enabling} which selectively masks spans at the granularities of words, n-grams, and sentences. We apply the masking function to each utterance multiple times to get up to 3 masked versions per utterance. Each utterance is constrained to have at least two masked spans. The spans are selected randomly for masking following~\citet{donahue-etal-2020-enabling}.

\noindent
\textbf{Infilling:} We extend the Infilling Language Model (ILM) from \citet{donahue-etal-2020-enabling} for dialogue response infilling (Figure~\ref{fig:ilm}). The ILM model is a GPT-2~\cite{Radford2019LanguageMA} based language model. For any piece of text $t$ with some spans masked with [blank] tokens, it is trained to predict the blanked spans in $t$ as a sequence generation problem. Each blank is infilled with an n-gram which can consist of one or more tokens.
For generating adversarial responses, infilling is done by conditioning on random contexts $C_{rand}$ instead of the original context $C$ to introduce various categories of errors (Table \ref{tab:categories}). For example in Figure~\ref{fig:ilm}, conditioning on a random context leads to the infilling of ``the marriage'' in the response, introducing error of type C-ent. For the context ``Did you stay your stay at our hotel?'' it generates a response ``I enjoyed at lot at the marriage''.
By corrupting the three types of utterances $R_g, U_c$ and $R_e$, this approach is able to introduce errors covering the 7 categories in Table~\ref{tab:categories}.

\noindent
\textbf{Preventing false negatives:}
Accidentally incorporating false negatives during training can lead to
the model learning to misclassify appropriate responses.
However due to the open-ended nature of dialogue responses, preventing generation of false negatives is not trivial.
In addition to conditioning on random contexts, we incorporate the following mechanisms during infilling to further reduce false negative generation:
\begin{itemize}[topsep=1pt,itemsep=0pt,partopsep=0pt, parsep=0pt,leftmargin=*]
\item \textit{Semantics of substitution}: We only select token substitutions which were not present in the tokens which were blanked. We also lower the generation probability of the blanked tokens' top 10 related words based on GloVe embedding~\cite{pennington2014glove} similarity by a factor of 100. This ensures that the blanks are not infilled by the originally blanked tokens or any related words.
\item \textit{Degree of substitution} - To ensure that the generated negative response is sufficiently different from the original utterance, we filter out the original utterance if the number of words in the utterance after stop-word removal is less than 2. We also filter a generated response if the difference in count of non stop-words between the original and generated response is less than 2.
\end{itemize}

\noindent
\textbf{Improving fluency:} The ILM model often generates responses with poor grammar or structure. To improve the fluency of the adversarial response sets, we first generate up to 4 different infilled variations of the masked original utterances, then score them using a GPT-2 based scorer named lm-scorer\footnote{\url{https://github.com/simonepri/lm-scorer}}. We then select the desired number of responses from this larger set. 

\begin{figure}[tb]
    \centering
    \includegraphics[width=0.46\textwidth]{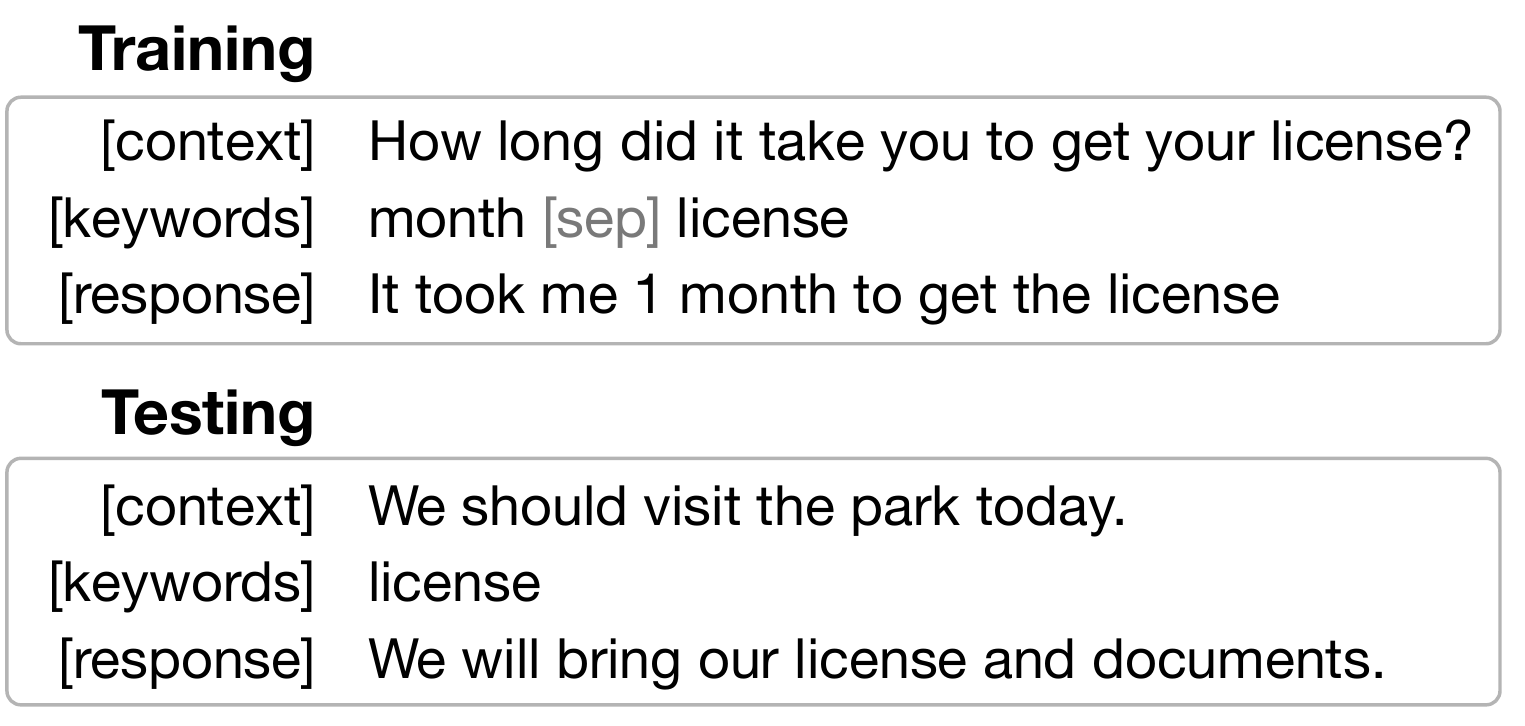}
    \caption{Keyword-guided approach for adversarial response generation. During training, the model learns to generate a response conditioned on its keywords and the correct context. During testing, it generates the response conditioned on a random context and keywords extracted from the correct context. The generated response thus shares content with the test context but does not directly address the context.}
    \label{fig:keymodel}
    \vspace{-1em}
\end{figure}

\subsection{Keyword-guided Approach}
This approach generates adversarial responses using keywords from the context as guidance, as shown in Figure~\ref{fig:keymodel}. 
The base generative architecture is a GPT-2 based dialogue model and it is trained to generate responses conditioned on the context and the response keywords. 
For adversarial response generation, the generation is conditioned on a random context ${C_{rand}}$ and keywords from the test context $C$. 
In Figure~\ref{fig:keymodel}, for the context ``How long did it take you to get your license?'' it generates a response ``We will bring our license and documents.''
To create the keyword set $K$ for a response, the model selects n number of keywords randomly from the set of all keywords extracted from the context $C$, where n is chosen randomly between 1 to 3 for every context. Keyword extraction is performed using Rake~\cite{rose2010automatic}. We call this model \textit{Key-context}. Since the generation is conditioned on keywords from context $C$, the generated response shares some content and semantics with the test context. However, since it is also conditioned on a random context ${C_{rand}}$, the generated response also incorporates entities, time expressions, speaker role, dialogue act, and other details based on ${C_{rand}}$. Since the generation model is not perfect, it also introduces errors related to fluency. Hence, the model is able to introduce errors covering the 7 categories in Table \ref{tab:categories}. 

\textit{Key-context} only uses keywords from the context to induce content similarity with the context. However, responses can have high content similarity due to the presence of similar concepts rather than just keywords. 
To introduce content similarity at concept level, we expand the keyword set $K$ with their top 10 most related words based on their GloVe embeddings. We use the gensim library\footnote{\url{https://radimrehurek.com/gensim/}} to find the most related words. For example, the related words for the keyword ``christmas'' are ``holidays'' and ``easter''. We replace a keyword in keyword set $K$ with one of its related words with a probability of $0.5$. We call this variant \textit{Key-sem}. 

\subsection{Classification Model}
\label{sec:class}
Our classification model architecture is based on the Speaker-Aware Bert (SA-Bert) model~\cite{10.1145/3340531.3412330}. Given a dialogue context $C = \{C_1, C_2, \ldots, C_h\}$ with $C_k$ denoting $k_{th}$ utterance in the context, a response $r$ and a label $y \in \{0,1\}$, the goal of the dialogue model $M$ is to learn a score $s(C,r)$ by minimizing cross-entropy loss function for the binary classification task. 
To calculate $s(C,r)$, $C$ and $r$ are concatenated, with a prepended [CLS] token. The output vector $\textbf{E}_{[CLS]}\in \mathbb{R}^{H}$ for the [CLS] token is used as the aggregated representation for the context-response pair classification. The final prediction is made as $\hat{y} = softmax(\textbf{W}\textbf{E}_{[CLS]})$, where $\textbf{W}\in \mathbb{R}^{2 \times H}$. SA-Bert model incorporates speaker information in two ways. First, an additional speaker embedding is added to the token representations which indicates the speaker’s identity for each utterance. Second, a [EOT] token is added at the end of each speaker turn. 
Before fine-tuning Bert model on the classification task, we first adapt Bert to the dataset by using the standard masked language model objective~\cite{devlin-etal-2019-bert}.

\section{Experiments}
\label{sec:experiments}
We test our approaches and baselines on dialogue classification, ranking and evaluation tasks.

\subsection{Training Details}
We use the base-uncased checkpoints for BERT~\cite{devlin-etal-2019-bert} and ELECTRA~\cite{Clark2020ELECTRAPT} from the Hugging Face transformers library~\cite{wolf-etal-2020-transformers}.
We trained the models with maximum sequence length of 128, maximum number of training epochs set to 3, Adam optimizer with initial learning rate of 5e-5 with linear decay, batch size of 60 per GPU on machines with 4 Nvidia 2080Ti GPUs. 
For generation, we use temperature of 0.9, nucleus sampling with $p$ equal to 0.9 and minimum length of 5.
We repeat each experiment three times (five times for BERT-based models) with different random seeds, use the validation split to select the best model, and report the mean metric values. 
Validation was done every 200 batches.

\subsection{Experimental Setup}

\subsubsection{Datasets}
We use two open-domain dialogue datasets: \textit{DailyDialog++}~\cite{sai2020improving} and PersonaChat~\cite{zhang-etal-2018-personalizing}.
{DailyDialog++} consists of 16900 dialogue contexts in train set, 1028 in validation set and 1142 in the test set. Each context contains 5 positive responses and 5 random negative responses. It also contains 5 adversarial responses per context collected through crowdsourcing where annotators were instructed to create negative responses with high content similarity with the context. A subset of 9259 out of the 16900 training contexts have 5 human-written adversarial negative responses.
It has two test sets, adversarial test set and random test set, based on the type of the negative response. 
\textit{PersonaChat} dataset~\cite{zhang-etal-2018-personalizing} is a corpus of human-human persona-conditioned conversations consisting of 8938 dialogues in the train set. We sample 2 random context-response pairs from each dialogue with a total of 17876 contexts for training. 
We prepend the persona utterances to the dialogue contexts in our experiments.
Since there is no human-created adversarial test set available for PersonaChat dataset, we construct an artificial adversarial dataset by randomly selecting an utterance from the dialog context and inserting it in the set of candidate responses following~\citet{jia-liang-2017-adversarial} and \citet{whang2020response}.
The adversarial test set for each context consists of the ground truth response, one utterance selected from the dialog context, and 8 random negative responses. The random test set consists of 9 random negative responses.

\noindent
\subsubsection{Metrics}
\label{sec:metrics}
For classification task, we report the accuracy  following~\cite{sai2020improving}.
For ranking task, we report standard ranking metrics - Recall $R_n@k$ and mean reciprocal rank (MRR). For DailyDialog++, $n$ is 6 in Recall as candidates consist of one positive response with 5 negative responses. For PersonaChat, $n$ is 10. 
For both classification and ranking tasks, we report results separately for the adversarial and the random test sets.

The dialogue evaluation task comprises of scoring or rating a response for its quality. For this task, we report the correlation of model scores with human provided ratings. We leverage the human ratings released by the following sources: 1) 600 ratings for response ``sensibility'' from~\cite{zhao2020multi} with inter-rater agreement $>$ 0.6 (Krippendorff’s $\alpha$ \cite{krippendorff2018content}). The responses consist of outputs from hierarchical recurrent encoder decoder (HRED) model with Attention~\cite{serban2016building} and Variational HRED model with attention~\cite{Serban_Sordoni_Lowe_Charlin_Pineau_Courville_Bengio_2017}; 2) 700 ratings for response quality from~\cite{zhao-etal-2020-designing}. The responses are from 6 different generative models - Seq-2-Seq~\cite{NIPS2014_a14ac55a}, attentional Seq-2-Seq, HRED, VHRED, GPT2-small, and GPT2-medium~\cite{Wolf2019TransferTransfoAT} with greedy decoding, ancestral sampling, and nucleus sampling based decoding~\cite{holtzman2019curious}. The inter-rater agreement is 0.815 (Krippendorff’s $\alpha$), and  3) Since the first two sources do not cover retrieval model outputs, we additionally collect quality ratings for 100 responses from a retrieval model's (Poly-Encoder~\cite{humeau2020poly}) selected responses and 100 human written responses with moderate inter-annotator agreement (Cohen's Kappa 0.45~\cite{cohen1968weighted}). All data points belong to the Dailydialog dataset and ratings are scaled between 0--1. By combining these sources we have a total of 1500 ratings for different context-response pairs.

\begin{table*}[tb]
\small
\centering
\begin{tabular}{l|l|ccc|ccc} 
\toprule
\textbf{Model}                 & \textbf{Approach}              & \multicolumn{3}{c|}{\textbf{Adversarial test set}}                                         & \multicolumn{3}{c}{\textbf{Random test set}}                                               \\
\multicolumn{1}{c|}{} & \multicolumn{1}{c|}{} & \multicolumn{1}{c}{Accuracy} & \multicolumn{1}{c}{R@1} & \multicolumn{1}{c|}{MRR} & \multicolumn{1}{c}{Accuracy} & \multicolumn{1}{c}{R@1} & \multicolumn{1}{c}{MRR}  \\ 
\hline
Poly-encoder          & Random                &     -                         &          0.684               &               0.806           &        -                      &         \textbf{0.849}                &         \textbf{0.914}                 \\
& Mask-and-fill  (Ours)         &    -                          &     0.758                    &        0.856                  &            -                  &        0.821                 &     0.897                     \\
& Key-sem (Ours)              &    -                          &             \textbf{0.788}             &                  \textbf{0.877}                 &            -                  &         0.828                &         0.902                 \\
\cline{2-8}
& Human                 &     -                         &      0.847                  &     0.913            &         -                     &     0.831                    &            0.902              \\ 
\hline

Electra               & Random                &     77.74                         &      0.915                   &          0.748                &       89.58                      &          0.957              &              \textbf{0.927}            \\
& Mask-and-fill (Ours)          &    \textbf{87.24}                          &        0.945                 &     \textbf{0.893}                    &        \textbf{89.61}                      &         \textbf{0.959}                &         \textbf{0.927}                 \\
& Key-sem  (Ours)             &        86.24                      &          \textbf{0.951}               &        0.881                  &              89.47                &        0.957                 &           0.924               \\
\cline{2-8}
& Human                 &        91.94                      &         0.984                &                0.967          &             87.95                 &    0.944                     &           0.911               \\ 
\hline

Bert                  & Random                &     77.82                         &                0.906         &           0.742               &    \textbf{89.34}                          &                \textbf{0.959}         &       \textbf{0.923}                   \\
& Semi-hard~\cite{li-etal-2019-sampling}             &     79.05                         &     0.913                   &                  0.756        &                 89.32              &        0.956                 &     \textbf{0.923}                     \\
& Token-subs~\cite{kryscinski-etal-2020-evaluating}             &            77.23                  &            0.901             &       0.783                   &             88.60                 &      0.950                   &    0.906                      \\
& BM25~\cite{Karpukhin2020DensePR}                  &      84.42                        &        0.936                 &       0.872                   &         87.68                     &                0.948         &       0.902                   \\

  & Mask-and-fill (Ours)           &   \textbf{87.45}                         &      \textbf{0.946}                   &        \textbf{0.904}                                        &  88.32                            &   0.951                      &             0.918                         \\
  & Key-context (Ours)          &            86.23                   &    0.939                          &       0.891        &    88.16      &       0.953 &           0.922                         \\
  & Key-sem (Ours)              &      87.02                                               &      0.944                   &       0.897                   &                  89.31          &      0.954                &    0.916                      \\
 \cline{2-8}
  & Human~\cite{sai2020improving}             &            91.22                      &       0.987                  &      0.973                                             &                       88.04       &  0.943                       &                                       0.901           \\
\bottomrule
\end{tabular}
\caption{Performance on classification and ranking tasks on DailyDialog++ test sets. Mask-and-fill and Key-sem approaches consistently perform the best across all model architectures compared to baselines on the Adversarial test set, just short of models trained with human created adversarial data.
Poly-encoder's accuracy is not available as it ranks candidates relative to each other.}
    \label{tab:ddmain}
    \vspace{-1.2em}
\end{table*}

\noindent
\subsubsection{Baselines}
We compare the following approaches of creating adversarial negative response sets. 

\noindent 
\textbf{Human}~\cite{sai2020improving} Human written adversarial responses. 

\noindent 
\textbf{Random} Responses sampled from random contexts.

\noindent
\textbf{Semi-hard}~\cite{li-etal-2019-sampling} Sampling scheme which selects samples from a batch based on their similarity scores with a margin of $\alpha$ from the positive response score. We perform static sampling and use Sentence-Bert~\cite{reimers-gurevych-2019-sentence} for semantic similarity calculation with $\alpha$ set to the recommended value of 0.07.

\noindent
\textbf{Token-subs}~\cite{kryscinski-etal-2020-evaluating} Training data is generated by applying a series of rule-based transformations on the positive responses. Transformations include pronoun, entity and number swapping, sentence negation and noise injection.

\noindent
\textbf{BM25} Top responses returned by BM25~\cite{DBLP:journals/ftir/RobertsonZ09} based on similarity with the context. Any ground truth response is removed from this response set if present by chance.
This baseline is inspired from~\citet{Karpukhin2020DensePR} and \citet{lin-etal-2020-world} and has shown strong performance in passage and response retrieval. 

\noindent 
\textbf{Mask-and-fill}  Our approach that infills utterances  conditioned on random contexts.

\noindent
\textbf{Key-context} Our approach that generates responses conditioned on test context keywords and random context history.

\noindent
\textbf{Key-sem} Our approach similar to Key-context which additionally conditions on words semantically related to the keywords in the context.

For each context, adversarial train sets are created by adding 5 random negative responses to the set of 5 negative responses created from the above approaches. If an approach create more than 5 responses, we randomly select 5 from them.

For dialogue evaluation, we compare the above approaches with BLEU, METEOR~\cite{banerjee2005meteor}, 
embedding based metrics 
SkipThought~\cite{10.5555/2969442.2969607},
Vec Extrema~\cite{forgues2014bootstrapping}, and RUBER~\cite{ruberbib} and BERTScore~\cite{bert-score}.

\noindent
\subsubsection{Models} We experiment with following architectures for ranking and evaluation models in our experiments: 1) Bert~\cite{devlin-etal-2019-bert}. We use the SA-Bert model~\cite{10.1145/3340531.3412330}, 2) Electra~\cite{Clark2020ELECTRAPT}, pre-trained with a replaced token detection objective and employs a generator-discriminator framework, and 3) Poly-encoders~\cite{humeau2020poly}, allows for fast real-time inference by precomputing each candidate response representation once, and then ranking candidate responses for retrieval by attending to the context. 




\subsection{Results and Discussion}
\label{sec:results}
In this section, we compare the performance of our approaches with the baselines on dialogue classification, ranking and evaluation tasks.

\skippingparagraph
\textbf{Performance on classification}
Our proposed approaches Mask-and-fill and Key-sem achieve the highest classification accuracy on the adversarial test set (Table \ref{tab:ddmain}), a few percentage short of the Human baseline.
The closest baseline is BM25 which has a gap of 3\% in accuracy compared to our approaches. 
Token-subs, which applies transformations on positive responses to corrupt them, does not fair well on this task.
This indicates that simple transformations do not provide good coverage of semantic variations present in the adversarial test responses.
Our approaches achieve similar performance across different model architectures, demonstrating their generalizability.
Unsurprisingly, the Human baseline performs strongly as the training and test data were created in the same manner and have similar distributions.
On the random test set, the performance of all approaches is either very close or lower than the Random baseline. 
Since the similarity between correct responses and the context is generally a lot higher than between random responses and the context in the random test set, Random baseline performs better since it associates coherence mostly with semantic similarity.
Finally, our analysis shows that all baselines tend to assign low scores to valid responses which do not address a context directly. For example, for the context ``Will you join us for the concert?'', if the response is ``It is supposed to rain this week.'', models assign it a low score. Such scenarios require understanding of social and commonsense related factors. We leave addressing this limitation to future work.

\begin{table}[tb]
\small
\centering
\begin{tabular}{l|ll|ll} 
\toprule
\textbf{Approach}    & \multicolumn{2}{c|}{\begin{tabular}[c]{@{}c@{}}\textbf{Adversarial}\\\textbf{test set} \end{tabular}} & \multicolumn{2}{c}{\begin{tabular}[c]{@{}c@{}}\textbf{Random}\\\textbf{test set}\end{tabular}}  \\
            &   R@1 & MRR     & R@1 & MRR         \\ 
\hline
Random               &   0.905  & 0.820      &        0.963    &  \textbf{0.914}           \\
Semi-hard            &   0.906  & 0.820                 &  \textbf{0.964 }  & 0.913               \\
Token-subs            &   0.895  & 0.825                 &  {0.958 }  & 0.901               \\
BM25           &  0.925    & 0.859     &          0.940    &  0.874            \\
Mask-and-fill (Ours)       &     \textbf{0.933}  & \textbf{0.871 }   &     0.952     &  0.890                 \\
Key-sem (Ours)    &    0.920 & 0.856       &            0.947   &  0.884              \\
\bottomrule
\end{tabular}
\caption{Performance on ranking task on PersonaChat dataset with Bert architecture. Our approaches perform better than all baselines on the adversarial test set.}
    \label{tab:personachat}
    \vspace{-1.0em}
\end{table}

\skippingparagraph
\textbf{Performance on ranking}
On the DailyDialog adversarial test set, Mask-and-fill and Key-sem approaches achieve the best Recall and MRR, closely followed by BM25 baselines (Table \ref{tab:ddmain}). 
The trends of the ranking metrics are similar to those observed for accuracy metrics. Our approaches perform better than the Human baseline on the random test set. 
On PersonaChat dataset, Mask-and-fill and Key-sem perform better than the baselines (Table~\ref{tab:personachat}), especially on the adversarial test set. This demonstrates the extensibility of our approach across datasets. Mask-and-fill performs better than Key-sem as the keyword sets contain a lot of keywords from the persona because of which responses have high content similarity with the persona rather than with the context.
The poor performance of the Random baseline provides evidence that training models using random negative candidates does not make the models robust against hard test cases during testing.
BM25 is a strong baseline for both datasets since retrieved responses also provide coverage over errors of various categories. 
However, retrieved response quality and diversity depends on the size of the retrieval pool. Furthermore, a stronger retrieval mechanism can lead to higher false negatives.
While the variation in BM25 response sets is constraint by the size of the dataset, and they provide lesser coverage over categories C-cont, C-strat and C-lang (Table~\ref{tab:categories}), our approaches have no such constraints.

\begingroup
\setlength\tabcolsep{2pt} 
\begin{table}[tb]
\small
\centering
\begin{tabular}{l|c|c} 
\toprule
\textbf{Approach}    & \multicolumn{1}{l|}{\textbf{Pearson}} & \multicolumn{1}{l}{\textbf{Spearman}}  \\ 
\hline
BLEU-2        &      0.046                         &     \underline{0.004}                          \\
METEOR~\cite{banerjee2005meteor}      &       0.081                        &     \underline{0.007}                          \\
SkipThought~\cite{10.5555/2969442.2969607}      &       0.059                        &     0.069                          \\
Vec Extrema~\cite{forgues2014bootstrapping}      &        0.157                       &      0.150                         \\
BERTScore~\cite{bert-score}   &          0.208                     &          0.198                     \\
RUBER~\cite{ruberbib}   &          0.253                     &          0.282                     \\
Random      &          0.296                     &          0.313                     \\
Semi-hard~\cite{li-etal-2019-sampling}  &          0.299                     &          0.315                      \\
BM25~\cite{Karpukhin2020DensePR}        &          0.310                     &          0.350                     \\
Token-subs~\cite{kryscinski-etal-2020-evaluating} &          0.324                     &          0.388                      \\
Mask-and-fill (Ours) &          0.338                     &          0.361                     \\
Key-sem (Ours)     &         \textbf{0.382}                      &          \textbf{0.401}                     \\
\hline
Human~\cite{sai2020improving}     &         0.348                      &          0.371                     \\
\specialrule{.8pt}{0pt}{1pt}
\end{tabular}
\vspace{-0.7em}
\caption{Comparison of approaches on dialogue evaluation. Trainable metrics are based on Bert architecture. For all entries except for the ones underlined, t-test \textit{p-value} $<0.05$. Mask-and-fill and Key-sem perform better than all baselines including the Human baseline.}
    \label{tab:evaluation}
        \vspace{-1.4em}

\end{table}
\endgroup

\newcommand\Tstrut{\rule{0pt}{2.6ex}}         
\newcommand\Bstrut{\rule[-0.9ex]{0pt}{0pt}}   

\begin{table*}[!htb]
\small
\centering
  \setlength\extrarowheight{-1pt}
\begin{tabular}{l|p{12.5cm}} 
\toprule

Context          & \begin{tabular}[c]{@{}l@{}}  A: Julia, will you be my wife?  \\ B: I'm sorry, Steven.  \\ C: Please, Julia, I have made proposal to you five times . I really want to share \\ \quad the rest of my life with you. \end{tabular}                                                                                         \\ \hline

Random           & \begin{tabular}[c]{@{}l@{}}  \rule{0pt}{0.8\normalbaselineskip}(1)  Yes of course it's a promise.  \\ (2) It's better to go somewhere else.  \\ (3) Let me first look at your work, how you have done it.  \\ (4) Being in love is a deep experience while having a crush is shallow.  \\ (5) Sometimes I don't understand, what is your problem? \end{tabular} 
\\ \hline
              
Mask-and-fill    & \begin{tabular}[c]{@{}l@{}}
\rule{0pt}{0.8\normalbaselineskip}(1)   You can't force me for to do that. They are designed for people of all ages and religions.  \\ (2) There you are. I'll have to make my own lunch!  \\ (3) I majored in economics. I really want i hope i can get some practical experience in life with you.  \\ (4) We will go to, and to meet some of the children who are visiting at school.   \\ (5) It takes time to learn. Bless you, baby! \end{tabular}  \\ \hline
Key-sem          & \begin{tabular}[c]{@{}l@{}} \rule{0pt}{0.8\normalbaselineskip}(1) And what about the potatoes? Steven, i don't know.  \\  (2) Sorry, there is no problem.  \\(3) Your wife didn't like it. Please don't tell me she is really interested in gardening.  \\ (4) I really want to go inside. It's really cold outside.  \\ (5) Really? I really want to pay a visit. I really want to spend the rest of my time enjoying this meal. \end{tabular}                                                                                                                                                                                                 \\ \hline
Human            & \begin{tabular}[c]{@{}l@{}} \rule{0pt}{0.8\normalbaselineskip}(1) I want to finish my home work by five and then I am going to take rest.  \\ (2) Follow these five tips, and you'll write a winning project proposal every time.  \\ (3) I met my wife a three to four times before the marriage.  \\ (4) Its difficult to live a life in a Dorze tribal area.  \\ (5) I shared a large number of ideas with the wedding planner. \end{tabular}                                                                                                                                                                                                  \\ 
\bottomrule
\end{tabular}
\caption{Sample adversarial responses from various approaches. Random responses are sampled from random dialogues. Human written responses are from the DailyDialog++ dataset. Mask-and-fill and Key-sem approaches create responses which are semantically related and yet inappropriate responses to the context. }
    \label{tab:examples}
    \vspace{-1.0em}
\end{table*}




\skippingparagraph 
\textbf{Performance on dialogue evaluation}
To study the performance of various approaches on real systems, we compare them on the task of Dialogue evaluation or scoring.
We measure the correlation between the scores predicted by the approaches in Table~\ref{tab:evaluation} with human provided ratings.
Reference based metrics like BLEU-2, METEOR, SkipThought and Vec Extrema achieve very low correlations, similar to findings reported in prior art~\cite{liu-etal-2016-evaluate, gupta-etal-2019-investigating}. BERTScore and RUBER achieve moderate correlation. Our approach Key-sem achieves the best correlations, followed by Mask-and-fill. BM25's performance is lower than that of our approaches, but it is higher than the Random and Semi-hard approaches. Although Token-subs did not achieve high performance on the classification and ranking tasks, it performs well on this task. This is likely because real model outputs contains more of the factual inconsistencies and contradictions that this approach captures, than what the adversarial test sets contain. 
Key-sem performs better than Mask-and-fill on evaluation since while Mask-and-fill only modifies utterances related to the context, Key-sem can freely generate more diverse adversarial responses for training.
Also, Key-sem achieves higher correlation than Human baseline. This may be because it is difficult for humans to create erroneous responses with distributions similar to the ones in model generated or selected responses, especially error types like C-speaker, C-strat and C-lang. In contrast, our approaches provide good coverage over all error types.

\begin{figure}[tb]
    \centering
    \vspace{-0.3pc}
    \includegraphics[width=0.4\textwidth]{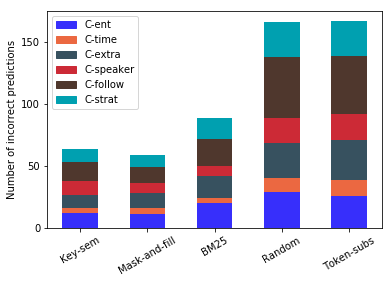}
    \vspace{-0.7em}
    \caption{Analysis of error types for different approaches on DailyDialog++ predictions. C-lang error type is not present in DailyDialog++. Mask-and-fill and Key-sem achieve a more uniform distribution over error categories compared to other approaches. }
    \label{fig:analyzeerror}
    \vspace{-1.0em}
\end{figure}

\skippingparagraph 
\textbf{Analysis of errors types}
We analyze the classification outputs of various approaches on the DailyDialog++ adversarial test set and report the types of misclassifications by each approach in Figure~\ref{fig:analyzeerror}. We first select a subset of test data where at least one of the approaches misclassifies the adversarial response as positive. We then manually categorize the types of errors presented in Table~\ref{tab:categories} for 200 randomly selected contexts from this subset. Each response can have multiple error types. C-follow and C-extra are the dominant error types which are misclassified by baselines Random, BM25 and Token-subs.
Key-sem and Mask-and-fill approaches achieve improvement in all error types compared to baselines and have a more uniform error distribution. While Key-sem performs better on C-extra, Mask-and-fill is better on C-follow and C-speaker. 

\skippingparagraph 
\textbf{Adversarial response examples}
We present sample responses from our approaches along with Random and Human baseline responses in Table~\ref{tab:examples}. 
Random approach generates responses which are easily distinguishable from ground truth responses. Mask-and-fill approach modifies either the ground truth response, utterances from the context or BM25 retrieved responses. It modifies these utterances to introduce corruptions such as non-contextual tokens, extraneous entities, incorrect time expressions, affective words or contradictions which makes the response either inappropriate or incoherent to the context, but it remains topically similar to the context. 
In Key-sem the dialogue acts, some entities and other tokens of the generated response depend on a random context the response is conditioned on, which also makes the response inappropriate or incoherent to the context.
\section{Related Work}
\label{related}
Dialogue response ranking and evaluation are important tasks in dialogue domain because even the recent large pretrained-language model based architectures~\cite{zhang2019dialogpt,humeau2020poly,Adiwardana2020TowardsAH,roller2020recipes,gupta-etal-2021-controlling} have been shown to be susceptible to creating inconsistent, ungrammatical and incoherent responses~\cite{roller2020recipes}. 
Traditional word-overlap based metrics like BLEU have been shown to be ineffective for dialogue response scoring~\cite{liu-etal-2016-evaluate, gupta-etal-2019-investigating}. Recently trainable metrics such as ADEM~\cite{lowe-etal-2017-towards}, RUBER~\cite{ghazarian-etal-2019-better} and USR~\cite{mehri-eskenazi-2020-usr} have been proposed for these tasks. However, since they are trained using negative samples obtained from random contexts, they are also prone to the spurious pattern of content similarity. 

Adversarial or counterfactual data creation techniques have been proposed for applications such as evaluation~\cite{gardner-etal-2020-evaluating, madaan2020generate},
attacks~\cite{ ebrahimi-etal-2018-hotflip, wallace-etal-2019-universal, jin2020bert}, explanations~\cite{goodwin-etal-2020-probing, ross2020explaining} or training models to be robust against spurious patterns and biases~\cite{10.1145/3306618.3317950,huang-etal-2020-reducing}. 
Adversarial examples are crafted through operations such as adding noisy characters~\cite{ebrahimi-etal-2018-hotflip,pruthi-etal-2019-combating}, paraphrasing~\cite{iyyer-etal-2018-adversarial}, replacing with synonyms~\cite{alzantot-etal-2018-generating,jin2020bert}, rule based token-level transformations~\cite{kryscinski-etal-2020-evaluating}, or inserting words relevant to the context~\cite{zhang-etal-2019-generating-fluent}.
While these approaches are optimized to change the predictions of a target model by perturbing the inputs, our approaches are more general and are not optimized towards any target model.
Polyjuice~\cite{wu2021polyjuice} and FactCC~\cite{kryscinski-etal-2020-evaluating} proposed approaches for model-agnostic general-purpose counterfactual generation. 
These approaches change the model's prediction by creating small edits through substitutions and insertions to the inputs. 
They are not applicable to our setting where we aim to flip the gold label, that is, convert a valid response to an adversarial response, while the model prediction should ideally remain the same to create hard training examples. Furthermore small perturbations do not provide good coverage over the adversarial response space and can create false negative responses.
Adversarial semantic collisions~\cite{song-etal-2020-adversarial} aims to generate texts that are semantically unrelated but judged as similar by NLP models to expose model vulnerabilities.
However, the outputs which are unrelated to the context are not useful for adversarial training as they are easy to classify. 

Finally, negative sampling strategies have also been studied for creating hard negative samples in context of visual embeddings~\cite{faghri2018vse++, Guo_Zhai_Yuan_Liu_Wang_2018}, knowledge graphs~\cite{kotnis2017analysis}, document retrieval~\cite{saeidi-etal-2017-effect, Karpukhin2020DensePR} and response retrieval~\cite{li-etal-2019-sampling, lin-etal-2020-world}. 
In this work we compare and build upon past work and are the first to propose generative approaches for adversarial negative response creation in dialogue.

\section{Conclusion}
This paper introduces approaches for synthesizing adversarial negative responses for training more robust dialogue response ranking and evaluation models. To synthesize a rich and comprehensive set of responses, we present and analyze categories of errors which affect the models. Our proposed approaches do not require any manual annotation and achieve high performance in dialogue classification, ranking and evaluation tasks across two datasets. 
These results demonstrate the promise of synthetic negative examples for improving open domain dialogue.
Future work, we will explore synthesizing adversarial test sets and methods for finer grained, controlled adversarial response generation.

\section*{Acknowledgements}
We thank Amy Pavel, Alissa Ostapenko, Rishabh Joshi, Artidoro Pagnoni and the anonymous  reviewers
for providing valuable feedback.
This work was funded by the Defense Advanced Research Planning Agency
(DARPA) under DARPA Grant N6600198-18908, and the National Science
Foundation under Awards No.~IIS1816012 and IIS2007960. Any opinions, findings, and conclusions or recommendations expressed in this material are those of the authors and do not necessarily reflect the views of the funding agencies.

\bibliographystyle{acl_natbib}
\bibliography{anthology,acl2021}

\clearpage

\appendix

\section{Quality of negative candidates}
We perform a human evaluation experiment to test the number of false negative responses created by the different approaches. Three in-house annotators were asked to go through the set of 5 adversarial negative responses from 5 different approaches for 100 randomly selected contexts. They were instructed to report the number of responses which are appropriate responses for the context, which in this case is the number of false negatives. After annotating separately, annotators finally discussed the responses marked as appropriate and aggregated the results. We observe that Human baselines responses had 2, Random baseline had 5, Mask-and-fill had 3, Key-sem had 4 and BM25 had 10 false negative responses in the set of 500 responses (100 contexts, with 5 adversarial responses each). This shows that our approaches do not generate high number of false negatives. BM25 on the other hand leads to a relatively higher number of false negatives which can impede the learning process of the models.

\section{Experiments with Masking}
\label{sec:maskingtype}
We experiment with two procedures for masking in the Mask-and-fill approach: 1) Random masking, which masks contiguous chunks of tokens some probability p. We leverage the masking function from \cite{donahue-etal-2020-enabling} which can selectively mask spans at the granularities of words, n-grams, and sentences. 2) Importance masking, which keeps the most important tokens in a response relevant to the context and masks the rest.
For Importance masking, we leverage the matching model from \cite{cai-etal-2019-retrieval} which is trained to estimate the sequence-level quality $s(q, r)$ of a response $r$ for a given query $q$. They decompose the sequence level matching score between a context and a response into a set of token-level scores as follows:
$$
\begin{aligned}
s(q, r) &=\mathbf{x}_{q}^{T} W^{s} \mathbf{x}_{r} \\
&=\mathbf{x}_{q}^{T} W^{s} \sum_{k=1}^{m} \omega_{k}\left(\mathbf{r}_{k}+\mathbf{e}_{r_{k}}\right) \\
&=\sum_{k=1}^{m} \omega_{k} \mathbf{x}_{q}^{T} W^{s}\left(\mathbf{r}_{k}+\mathbf{e}_{r_{k}}\right)
&=\sum_{k=1}^{m} \omega_{k} s_{k}
\end{aligned}
$$
where $s_{k}=\mathbf{x}_{q}^{T} W^{s}\left(\mathbf{r}_{k}+\mathbf{e}_{r_{k}}\right)$, and $x_r$ is the weighted sum of a Bert Transformer encoder outputs $r_{k}$ as well as their initial vector representations $e_{k}$.
The importance of each response token $k$ to the context is estimated by $s_{k}$. We mask out any token with importance weight $\omega_{k}$ less than the average $\omega$ and only retain tokens highly relevant to the context following \citet{cai-etal-2019-retrieval}. In our initial experiments we found that the Importance masking procedure lead to worse performance than Random masking. The adversarial test set accuracy on DailyDialog adversarial test set was 85.43\% compared to the 87.45\% accuracy using Random masking. Our analyses showed that Importance masking masked out about 50\% of the response tokens, and the infills generated by the ILM model were mostly poor in fluency as the number of masked tokens was high. We therefore finally used Random masking for Mask-and-fill.

\begin{table*}[!htb]
\small
\centering
\begin{tabular}{l|p{12.5cm}} 
\toprule
Context or Model & Utterances                                                                                                                      \\ 
\hline
Context          & \begin{tabular}[c]{@{}l@{}}A: OK . Now I'll put the dishes in the sink.~\\B: Thank you . I'll wash them.~\\C: OK . Then I will clean the table.\end{tabular}                                                                                                                                                                                                                                                                                                                                                                                                                                  \\
 \hline
Random           & \begin{tabular}[c]{@{}l@{}}(1) May I have your name and number, Sir?  \\(2) Then I hope to get the other documents by end of the day.  \\(3) She is very experienced in that area, including all the years in college  \\(4) I am in Computer Science department.  \\(5) Sure, you can talk to him. \end{tabular}                                                                                                                                                                                                                                                                             \\  \hline
Mask-and-fill    & \begin{tabular}[c]{@{}l@{}} (1)  Ok. Thank you. But, what are you going to do with him?  \\ (2) Uh, it's time to wake up. I will clean it up and then go to bed.  \\ (3) Oh, thank you. You have helped a lot.  \\ (4) Don't, thank you very much for saying it like that. Only in your opinion do you have to wear \\ proper clothes.   \\ (5) Yes, thank you! But, i am not satisfied with the work you've done. \end{tabular}                                                                                                                                                                                     \\  \hline
Key-sem          & \begin{tabular}[c]{@{}l@{}} (1)  Sorry, he didn't put the dishes on the table.  \\ (2) Ok. We'll clean up the room now. I can walk you through the process.  \\ (3) Don't forget to wash the dishes and put away the clothes.  \\ (4) In my field, i put on quite a few weight -bearing exercise in order to improve my lung capacity.  \\ (5) Thank you for your understanding. What are your recipes for tableware? \end{tabular}                                                                                                                                                                              \\ \hline
Human            & \begin{tabular}[c]{@{}l@{}} (1)  I just now saw the news that the boat was sinking due to heavy goods.  \\ (2) I want to thank my friend because he helped me to wash my dress at school camp.  \\ (3) Nowadays, table fans are getting very cheap online.  \\ (4) I know that using a facial scrub can make your skin look beautiful, clean and soft.  \\ (5) I gifted a sink to my friend for his house warming ceremony. \end{tabular}                                                                                                                                                                        \\  
\hline \hline
Context          & \begin{tabular}[c]{@{}l@{}}A: Can you tell me what's my responsibility in this position?  \\ B: Yes, of course . You would be responsible for the development of software products.  \\ C: I see . This is my advantage. \end{tabular}                                                                                                                                                                                                                                                                                                                                                                 \\ \hline
Random           & \begin{tabular}[c]{@{}l@{}} (1)  Okay! That sounds great to me.  \\ (2) Well! How much will it cost per kg?  \\ (3) Well! You can pay it on monthly or yearly basis, it is upto you.  \\ (4) I usually spend those days with my family and it is quite fun you see.  \\ (5) What type of games do you like to play? \end{tabular}                                                                                                                                                                                                                                                                                \\ \hline
Mask-and-fill    & \begin{tabular}[c]{@{}l@{}} (1) Yes. Maybe he is just looking for some publicity. You are responsible, too.   \\(2) I see. Then we will all get on our own.  \\(3) That's nice. And i would be willing to take them for that.  \\(4) You also have to work on the meetings to be more focused. I need to add some training.  \\(5)  What kind of software do they use now? \end{tabular}                                                                                                                                                                                                                      \\ \hline
Key-sem          & \begin{tabular}[c]{@{}l@{}}  (1)  Let me see, in your brochure, what kind of promotion you're promising?  \\(2) Tell me about it. What do you think? Will you marry her?  \\(3) Of course. Of course there are many things online. Tell me about it.  \\(4) Yes, i appreciate your cooperation. The development of the l / c is our utmost priority.  \\(5) Thank you. I do want to get him a diamond ring. He's responsible for development of the etv. \end{tabular}                                                                                                                                        \\ \hline
Human            & \begin{tabular}[c]{@{}l@{}}  (1)  Of course, the museum is in the closing stage because of financial issues.  \\(2) I was searching on some websites for the junior engineer position to develop my knowledge\\ in the hardware field.  \\ (3) I see, is there any terms and condition that I have to sign for this position in your company?  \\(4) Of course, you must provide me the full details about our company's financial position by\\ today evening.  \\(5) Of course, My friend is very much interested to work in a software company. Can you give\\ him a chance in your company? \end{tabular}     
\\ 
\bottomrule
\end{tabular}
\caption{Outputs from different approaches for negative response set creation. Random responses are unrelated to the contexts. Mask-and-fill and Key-sem approaches create responses which are highly similar to the content of the contexts, and hence the model needs to learn factors important for response coherence and appropriateness such as presence of correct entities, time expressions, strategies and others.}
    \label{tab:examples2}
\end{table*}

\section{Sample Model Generated Responses}
In continuation of sample responses presented in Table~\ref{tab:examples} of the main paper, we present some more sample responses from different approaches along with Random and Human baseline responses in Table~\ref{tab:examples2}. 

\section{Additional Implementation Details}
For BM25 approach, we use the open source implementation from transformer rankers\footnote{https://github.com/Guzpenha/transformer\_rankers}. The DailyDialog++ dataset contains 16900 dialogue contexts but only 9259 of those have adversarial negative responses for the Human baseline. For the results reported in Table~\ref{tab:evaluation}, all approaches from Random and below use the Bert architecture and trained using DailyDialog domain data. Additionally, RUBER is also trained on the DailyDialog++ dataset. The approaches above Random in the table do not require training. Each approach predicts a score for the set of 1500 responses created using a set of generative and retrieval models as detailed in section~\ref{sec:metrics}.
Sentence-Bert used in Semi-hard sampling scheme is fine-tuned on the datasets used in this paper.

For the Mask-and-fill approach, the model takes in the following sequence of inputs: \{[context] $C_1$ [eot],.., [eot] $C_h$ [response] $r$-with-blanks [infill] $B_1$ [answer],.., $B_l$ [answer]\}, where ${C}_{c=1}^{h}$ represents a context with h utterances, $r$ the response and ${B}_{b=1}^{l}$ are the tokens blanked in the response. [eot] is used to indicate end of turn. To generate a set of 5 adversarial responses in the Mask-and-fill approach, we first create 4 masked versions of every utterance related to the context ($R_g, U_c$ and $R_e$). ILM model then generates 4 infills per masked utterance. Thus each utterance gets 16 different modified versions. All these modified utterances are then ranked using the lm-scorer library and we select the top 5. BM25 similarity is used to create the retrieved response set. 

For the Keyword-guided approaches, the model is given as input the context $C$, keywords from the ground truth response $K$, and the ground truth response $r$ as shown in Figure~\ref{fig:keymodel}.
Specifically, the model takes in the following sequence of inputs - \{[context] $C_1$ [eot],.., [eot] $C_h$ [keywords] $K_1$ [sep],..,[sep] $K_n$ [response] $r$\}. For both approaches during training, positive responses and negative responses are interleaved, i.e. each positive response is followed by one random and one adversarial response.

\end{document}